\documentclass[conference,10pt]{IEEEtran}
\IEEEoverridecommandlockouts
\usepackage{cite}
\usepackage{amsmath,amssymb,amsfonts}
\usepackage{algorithmic}
\usepackage{graphicx}
\usepackage{textcomp}
\usepackage{xcolor}
\def\BibTeX{{\rm B\kern-.05em{\sc i\kern-.025em b}\kern-.08em
    T\kern-.1667em\lower.7ex\hbox{E}\kern-.125emX}}
\usepackage{latexsym}
\usepackage{amssymb}
\usepackage{amsmath}
\usepackage{amsthm}
\usepackage{booktabs}
\usepackage{enumitem}
\usepackage{graphicx}
\usepackage{color}
\usepackage{xspace}
\usepackage{dutchcal}
\usepackage[font=small]{caption}
\usepackage{array}

\renewcommand{\arraystretch}{1.2}
\newcommand{\projectname}{HyperKAN\xspace}

\DeclareRobustCommand*{\IEEEauthorrefmark}[1]{%
    \raisebox{0pt}[0pt][0pt]{\textsuperscript{\footnotesize\ensuremath{#1}}}}
    
\begin{document}

% \topmargin=0mm
% \begin{frontmatter}
% \paperid{981} 
% \title{Enhancing Hypergraph Neural Networks Through Structural Fusion for Improved Expressiveness}
\title{HyperKAN: Hypergraph Representation Learning with Kolmogorov-Arnold Networks}
\author{
    % \IEEEauthorblockN{Xiangfei Fang\IEEEauthorrefmark{1}\IEEEauthorrefmark{2}, Heng Zhang\IEEEauthorrefmark{1}, Chen Zhao\IEEEauthorrefmark{1}}
        
    % \IEEEauthorblockA{\IEEEauthorrefmark{1}Institute of Software, Chinese Academy of Sciences,}
    % \IEEEauthorblockA{\IEEEauthorrefmark{2}University of Chinese Academy of Sciences}
    % \IEEEauthorblockA{\IEEEauthorrefmark{3}North University of China}
   \IEEEauthorblockN{Xiangfei Fang\IEEEauthorrefmark{1,2}, Boying Wang\IEEEauthorrefmark{3*}\thanks{*Corresponding Author.}\thanks{This work was supported in part by National Natural Science Foundation of China(No. 62002350), in part by Youth Innovation Promotion Association CAS(No. 2023120), and in part by the Youth Project of Applied Basic Research Project of Shanxi Province (entitled "Research on General Object
Detection Algorithms Driven by Multi-Source Imbalanced Data").}, Chengying Huan\IEEEauthorrefmark{4}, Shaonan Ma\IEEEauthorrefmark{5},  Heng Zhang\IEEEauthorrefmark{1}, Chen Zhao\IEEEauthorrefmark{1}}
    
    \IEEEauthorblockA{\IEEEauthorrefmark{1}Institute of Software, Chinese Academy of Sciences,\IEEEauthorrefmark{2}University of Chinese Academy of Sciences,}
    \IEEEauthorblockA{\IEEEauthorrefmark{3}North University of China,\IEEEauthorrefmark{4}Nanjing University,\IEEEauthorrefmark{5}Qiyuan Lab}
}

\maketitle
\begin{abstract}
 Hypergraph representation learning has garnered increasing attention across various domains due to its capability to model high-order relationships. Traditional methods often rely on hypergraph neural networks (HNNs) employing message-passing mechanisms to aggregate vertex and hyperedge features. However, these methods are constrained by their dependence on hypergraph topology, leading to the challenge of imbalanced information aggregation, where high-degree vertices tend to aggregate redundant features, while low-degree vertices often struggle to capture sufficient structural features.

To overcome the above challenges, we introduce \projectname, a novel framework for hypergraph representation learning that transcends the limitations of message-passing techniques. \projectname begins by encoding features for each vertex and then leverages Kolmogorov-Arnold Networks (KANs) to capture complex nonlinear relationships. By adjusting structural features based on similarity, our approach generates refined vertex representations that effectively addresses the challenge of imbalanced information aggregation. Experiments conducted on the real-world datasets demonstrate that \projectname significantly outperforms state-of-the-art HNN methods, achieving nearly a 9\% performance improvement on the Senate dataset.
\end{abstract}
% \end{frontmatter}

\section{Introduction}
A hypergraph, as a generalized form of a graph where edges (hyperedges) can connect more than two vertices, excels in representing complex relationships and is widely applied in real-world scenarios such as social networks \cite{li2013link,meng2024link,guan2023sparse,10887924}, biological systems \cite{feng2021hypergraph,jin2023general,lugo2021classification}, and computer vision \cite{shi2020point,pradhyumna2021graph,han2022vision}, where data often exhibit multi-way relations that need to be effectively modeled. To better capture these higher-order relationships and improve the performance of machine learning models on complex data, numerous methods for hypergraph representation learning have been extensively explored \cite{antelmi2023survey,liu2024hypergraph,kim2024hypeboy}.

Most of these methods \cite{huang2021unignn,chien2021you,dong2020hnhn} focus on enhancing hypergraph representation learning by designing novel hypergraph neural network (HNN) architectures. These approaches employ operators such as convolution, attention, and diffusion models to facilitate more efficient vertex feature learning. A prevalent strategy is to refine the two-stage message-passing mechanism, which involves first aggregating features from vertices to hyperedges and then updating features from hyperedges back to vertices. Through this iterative message-passing mechanism, vertices in the hypergraph can gather essential information from their neighbors and the surrounding structure, thereby optimizing representation learning for various tasks.

However, the message-passing mechanism is fundamentally constrained by its reliance on the hypergraph's original topology, leading to the challenge of imbalanced information aggregation. This issue arises from the degree heterogeneity in hypergraphs and the mechanism's limited focus on vertex-to-hyperedge and hyperedge-to-vertex interactions. As a result, vertices with few neighbors may receive insufficient information, while vertices with many neighbors can be overloaded with redundant information. This imbalance in information flow adversely affects the learning process, ultimately resulting in suboptimal performance in downstream tasks.

To address this challenge, we propose a novel framework for hypergraph representation learning, termed \projectname. Unlike traditional methods based on message-passing mechanisms, \projectname initially extracts feature encodings for each vertex from the hypergraph and then utilizes these encoded features as inputs to Kolmogorov-Arnold Networks (KANs) for effective vertex representation learning.
We start by computing each vertex's feature vector as a weighted sum of its initial features and those of its neighbors.  
To tackle the issue of imbalanced information aggregation, we first adjust the structural features using the similarity features which can reduce redundant features for high-degree vertices and enrich the information for low-degree vertices. We then leverage the powerful dynamic adjustment and adaptability capabilities of KANs to effectively capture complex, nonlinear relationships and further extract higher-order features from these adjusted structural features. This approach further reduces redundancy for high-degree vertices and enhances information acquisition for low-degree vertices, further addressing the challenge of imbalanced information aggregation.

We conduct experiments on four real-world datasets and compare its performance with state-of-the-art HNNs to validate the effectiveness of \projectname. The experimental results show that \projectname significantly outperforms existing methods, achieving nearly a 9\% improvement on the Senate dataset.
% \vspace{-1em}

\begin{figure*}
    \centering
\includegraphics[width=0.95\textwidth]{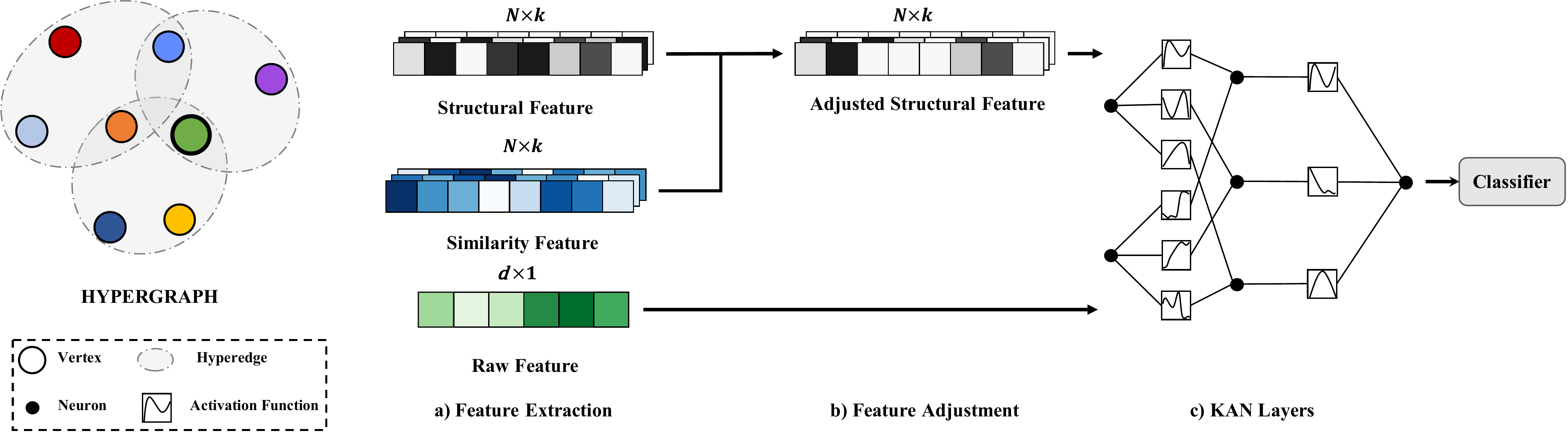}
% \vspace{0.1em}
% \vspace{0.5em}
\captionsetup{justification=raggedright}
    \caption{Architecture of the \projectname model where $N$ is the number of vertices, $k$ is the hop number, and $d$ is the dimension of the vertex feature vector. The model comprises three key components: a) the Feature Extraction module, which extracts structural features from the hypergraph structure and similarity features from the raw vertex features; b) the Feature Adjustment module for structural feature adjustment utilizing the similarity features; and c) the KAN layers for further encoding and feature extraction.}
    % \vspace{-0.5em}
    \label{fig:enter-label}
\end{figure*}

\section{Background and Related Work}
\label{gen_inst}
\textbf{Hypergraph.} A hypergraph $\mathcal{H} = (\mathcal{V},\mathcal{E})$ consists of a set of vertices $\mathcal{V}$ and hyperedges $\mathcal{E}$, where each hyperedge $e \in \mathcal{E}$ is a subset of $\mathcal{V}$ and can connect multiple vertices. This structure allows hypergraphs to capture complex, higher-order relationships. The incidence matrix $\mathbf{H}$, where $\mathbf{H}_{ij} = 1$ if vertex $i$ is in hyperedge $j$ and $\mathbf{H}_{ij} = 0$ otherwise, is used to represent these connections efficiently.

\textbf{Hypergraph Neural Networks.}
Recent research has focused on developing hypergraph neural networks (HNNs) to efficiently learn representations from hypergraphs. Some approaches convert hypergraphs into standard graphs and apply message-passing techniques for convolution operations on these converted structures \cite{yadati2019hypergcn,sun2008hypergraph,yang2022semi,yan2024hypergraph}. Other studies have proposed new HNN architectures based on a two-stage message-passing mechanism, such as HGNN \cite{feng2019hypergraph}, UniGNN \cite{huang2021unignn}, AllSet \cite{chien2021you}, and others \cite{arya2020hypersage,tudisco2021nonlinear,bai2021hypergraph,wang2022equivariant,duta2023sheaf}. Despite their innovations, these methods are inherently limited by the hypergraph or the converted graph structure, as they typically only capture information from immediate neighbors. This restricts the potential of hypergraph representation learning to fully utilize the rich, higher-order relationships present in the data.

\textbf{Kolmogorov-Arnold Networks.}
Kolmogorov-Arnold Networks (KANs) offer a novel approach to representation learning, particularly suited for modeling complex, nonlinear relationships. Unlike traditional neural networks like Multi-Layer Perceptrons (MLPs) that use fixed activation functions, KANs use learnable activation functions parameterized as splines, allowing them to adapt to diverse data patterns and enhance modeling flexibility and accuracy \cite{liu2024kan}. Recently, KANs have been applied to graph-structured data \cite{sa,de2024kolmogorov,kiamari2024gkan}, but their use for hypergraph-structured data remains unexplored. To the best of our knowledge, our work is the first to employ KANs for hypergraph representation learning.

\section{Proposed \projectname Architecture}
We introduce a novel architecture, named \textbf{\projectname}, for hypergraph representation learning, as illustrated in Fig.~\ref{fig:enter-label}.
\projectname starts by leveraging the connectivity and similarity properties of vertices in the hypergraph to extract two key types of features: structural features and similarity features (Section \ref{4.1}). To address the challenge of imbalanced information aggregation, \projectname applies an adjusting process on the structural features using similarity (Section \ref{4.2}). The adjusted structural features are then encoded and processed through Kolmogorov-Arnold Network (KAN) layers to capture deeper, more complex relationships (Section \ref{4.3}). Finally, the vertex classification layer generates the final classification result (Section \ref{4.4}).

\subsection{Hypergraph Feature Extraction Module}
\label{4.1}
\textbf{Structural feature.}
A vertex $v_i \in \mathcal{V}$ gathers features from its neighboring vertices connected by hyperedges, which can be represented using the adjacency matrix of vertices, $A = \mathbf{H} \mathbf{H}^{\top}$. Building on this adjacency matrix, we extract $k$-hop neighbor features to leverage the hypergraph's structural properties, resulting in a comprehensive structural feature representation formulated as $h_i = \left[ A_i, A^2_i, A^3_i, \ldots, A^k_i \right]$, where $A_i$ represents the $i^{th}$ row of the adjacency matrix $A$, which corresponds to the direct neighbors of vertex $v_i$, $A^2_i$  refers to the $i^{th}$ row of $A^2$ indicating the 2-hop neighbors of $v_i$, and this pattern continues up to $A^k_i$ for the $k$-hop neighbors.

The structural feature vector $h_i$ encapsulates multi-hop neighborhood information for vertex $v_i$, providing a richer representation of the hypergraph's topology. Unlike the vertex's original features, these features derived from its neighbors encapsulate the structural context of the hypergraph, offering a more comprehensive understanding of relationships within it. This enriched representation enhances the model's ability to capture complex dependencies and interactions, which is crucial for subsequent classification tasks.

\textbf{Similarity feature.} To further enhance the representation of vertices in the hypergraph, we compute similarity features by evaluating how each vertex relates to all others. For this, we calculate a similarity score between each pair of vertices to form a graph similarity matrix $S \in \mathbb{R}^{\lvert \mathcal{V} \rvert \times \lvert \mathcal{V} \rvert}$, where each element $S(i,j)$ indicates the similarity score between vertices $v_i$ and $v_j$. To efficiently construct this matrix, we use cosine similarity. Let $X \in \mathbb{R}^{n \times d}$ be the feature matrix, where each row $X_i \in \mathbb{R}^d$ represents the feature vector of vertex $v_i$ in a $d$-dimensional space. The cosine similarity matrix $S \in \mathbb{R}^{n \times n}$ is defined to measure the similarity between pairs of vertices based on their feature vectors in $X$. Specifically, each entry $S_{ij}$ in matrix $S$ is computed as:

\begin{equation}
\begin{aligned}
\label{eq2}
S_{ij} = \frac{X_i \cdot X_j}{\|X_i\| \|X_j\|},
\end{aligned}
\end{equation}

where $X_i$ and $X_j$ are the feature vectors of vertices $v_i$ and $v_j$, respectively; $\cdot$ denotes the dot product, and $|\cdot|$ denotes the Euclidean norm. This cosine similarity matrix $S$ captures the pairwise similarities between vertices based on the angles between their feature vectors. The similarity features of a vertex $v_i$ are given by the row vector $s_i$ of matrix $S$, where $s_i \in \mathbb{R}^n$ contains the cosine similarity values between vertex $v_i$ and all other vertices. This representation effectively reflects vertex relationships using their raw feature information.

\subsection{Feature Adjustment Module}
\label{4.2}
 To address the challenge of imbalanced information aggregation, we employ an adjustment mechanism based on similarity features to refine the structural features. We design a filtering function $\mathcal{F}$ to adjust the structural feature $h_i$, reducing redundant neighbors for densely connected vertices while incorporating important multi-hop neighbors for sparsely connected ones. We use the feature adjustment of $A_i$ as an illustrative example.

If the $j^{th}$ element of $A_i$ is non-zero, it indicates a connection between vertex $v_i$ and vertex $v_j$. To mitigate redundancy from multiple neighboring vertices, we utilize the similarity features $S_i$ for adjustment. Specifically, if a vertex has more than $n$ neighbors, we retain only the top-$n$ neighbors with the highest similarity scores and remove those with lower similarity. For such cases, we set $A_{ij} = 0$ if $S_{ij} \notin \text{top-}n(S_i)$. After adjusting, a vertex will retain features from at most $n$ neighbors that are most similar to its own features.
 
Conversely, if a vertex has fewer than $m$ direct neighbors, we add extra neighbors to its structural feature based on the similarity features $S_i$. We select additional neighbors that have the most similar characteristics based on the similarity scores. For each vertex $v_i$, its similarity feature $S_i$ is used for adjustment as follows:
\begin{equation}
\begin{aligned}
\label{eq3}
\text{Mask}_{ij} = 
\begin{cases} 
1, & \text{if } A_{ij} = 0 \\
0, & \text{otherwise}
\end{cases}, \\
\end{aligned}
\end{equation}

We then set $A_{ij} = 1$ if $S_{ij} \in \text{Top-}m(S_i \cdot \text{Mask}_i)$ where $\cdot$ represents the dot product. 

The above processes eliminate redundant structural features for high-degree vertices and add necessary hypergraph structural information for low-degree vertices. Such a balanced and informative set of neighbors is crucial for capturing the underlying hypergraph structure effectively.
To formalize these operations, we denote these two aforementioned filtering functions as $\mathcal{F}_1$ and $\mathcal{F}_2$ separately. With these functions in place, the adjusted structural features can be represented as:
\begin{equation}
\begin{aligned}
\label{eq4}
\hat{\mathbf{A}}_i = \frac{\hat{A_i}}{d_i},
\end{aligned}
\end{equation}

where $\hat{A_i} = \mathcal{F}1(\mathcal{F}2(A_i))$ and $d_i$ is defined as $d_{i} = \sum\limits_{j} \hat{A}_{ij}$. Similarly, the remaining components of $h_i$ are adjusted, resulting in a refined structural feature denoted as $\hat{h}_i = \left[ \hat{\mathbf{A}}_i, \hat{\mathbf{A}}^2_i, \hat{\mathbf{A}}^3_i, \ldots, \hat{\mathbf{A}}^k_i \right]$.

\subsection{KAN Layers}
\label{4.3}
Building upon the adjusted structural features defined above, we perform an aggregation with the raw features $x$ of each vertex to enrich the representation by combining both structural and attribute information. The aggregation process for a vertex $v_i$ is defined as:
\begin{equation}
\begin{aligned}
\label{eq5}
e_i = Aggregate(\sum_{j=1}^{k} (\hat{h}_{ij} x_i), x_i).
\end{aligned}
\end{equation}
 In this paper, we define the aggregation function as the weighted sum of vectors. Subsequently, we map the aggregated vector to a $d$-dimensional feature space, represented as $z_i = Embed(e_i) \in \mathbf{R}^{d \times 1}$.
%  \begin{equation}
% \begin{aligned}
% \label{eq6}
% z_i = Embed(e_i) \in \mathbf{R}^{d \times 1}.
% \end{aligned}
% \end{equation}
The features of all vertices are organized into a matrix form $Z^{0} = [z_1, z_2, \ldots, z_{k}] \in \mathbb{R}^{k \times d}$, where $k$ is the count of vertices. The matrix $Z^0$ is then fed into the KAN layers as input to update the vertex representations across multiple layers.
After $l$ KAN layers, the initial feature $Z^{0}$ is transformed into a higher-level representation $Z^{l}$. This process can be mathematically described as:

\begin{equation}
\begin{aligned}
\label{eq7}
Z^{l} = \left( \Phi_{L-1} \circ \cdots \circ \Phi_1 \circ \Phi_0 \right) Z^{0},
\end{aligned}
\end{equation}

 where $\Phi_L$ represents a KAN layer and is given by:
 
\begin{equation}
\begin{aligned}
\label{eq8}
\Phi_L =
\left(
\setlength\arraycolsep{3pt}
\begin{array}{cccc}
\phi_{l,1,1}(\cdot) & \phi_{l,1,2}(\cdot) & \cdots & \phi_{l,1,n_l}(\cdot) \\
\phi_{l,2,1}(\cdot) & \phi_{l,2,2}(\cdot) & \cdots & \phi_{l,2,n_l}(\cdot) \\
\vdots & \vdots & \ddots & \vdots \\
\phi_{l,n_l+1,1}(\cdot) & \phi_{l,n_l+1,2}(\cdot) & \cdots & \phi_{l,n_l+1,n_l}(\cdot) \\
\end{array}
\right),
\end{aligned}
\end{equation}

where each $\phi_{l,i,j}$ is the activation function that connects $i^{th}$ neuron in the $l^{th}$ layer and the $j^{th}$ neuron in the $l+1^{th}$ layer.

\subsection{Loss Function}
\label{4.4}
For the vertex classification task, we train our model using the cross-entropy loss function, which is defined as:
\begin{equation}
\begin{aligned}
\label{eq9}
\mathcal{L} = -\frac{1}{N} \sum_{i=1}^N \sum_{c=1}^C \textbf{y}_{i,c} \log(\hat{\textbf{y}}_{i,c}),
\end{aligned}
\end{equation}

where $N$ is the total number of labeled vertices,  $C$ is the number of classes, $\textbf{y}_{i,c}$ represents the ground-truth label of $i$-th sampled vertex, and $\hat{\textbf{y}}_{i,c}$ represents the prediction value.

\section{Evaluation}
\subsection{Experimental settings}
\textbf{Datasets.} We utilize four datasets to evaluate our model's performance on the node classification tasks. The first dataset is a hypergraph-based citation network, Citeseer \cite{yadati2019hypergcn}. Another dataset is an adaptation of the Senate data \cite{fowler2006legislative}, as proposed in \cite{chodrow2021hypergraph}. The final two datasets, ModelNet40 \cite{wu20153d} and NTU2012 \cite{chen2003visual}, are used for visual object classification tasks.
Table \ref{GraphData Table} provides a comprehensive overview of these datasets.

\textbf{Baselines.} The performance of the model is evaluated through classification tasks on hypergraph vertices. We compare our approach against several baseline models, including HGNN \cite{feng2019hypergraph}, HyperGCN \cite{yadati2019hypergcn}, UniGNN \cite{huang2021unignn}, HNHN \cite{dong2020hnhn}, HCHA \cite{bai2021hypergraph}, AllSet \cite{chien2021you}, $\mathrm{HGNN}^{+}$ \cite{9795251}, and EDHNN \cite{wang2022equivariant}. All these baselines have open-source implementations available, ensuring that the results are reproducible.

\renewcommand{\arraystretch}{1.1}
\begin{table}[ht!]
\caption{Dataset statistics in our evaluation.}  
  \label{GraphData Table}
  \centering
  \resizebox{0.85\linewidth}{!}{ 
  \begin{tabular}{ccccc}
    \toprule
    % \multicolumn{2}{c}{Part}                   \\
    % \cmidrule(r){1-2}
    Dataset  & \#Vertices & \#Edges &\#Vector Dimensions & \#Classes \\
    \midrule
    % Cocitation Cora & 2708  & 1579 & 1433 & 7   \\
    % Cocitation Pubmed & 19717 & 7963 & 500 & 3 \\
    Citeseer & 3327 & 9464 & 500 & 6 \\
    % Coauthorship Cora  & 2708  & 1072 & 1433 & 7\\
    % House       & 1290    & 340     & 100   & 2 \\
    Senate       & 282   & 315     & 100   & 2 \\
    ModelNet40    & 12311 & 12311   & 100   & 40 \\
    NTU2012    & 2012 & 2012  & 100   & 67 \\
    \bottomrule
  \end{tabular}
}
\end{table}

\textbf{Experiment Settings.} 
We adhere to the standard practice of partitioning each dataset into training, validation, and testing sets, following a ratio of $2:1:1$. For optimization, we use the AdaGrad algorithm \cite{kingma2014adam}. To ensure the robustness and reliability of our results, each model is trained and validated ten times with different configurations. The evaluation metric used is prediction accuracy. All experiments are conducted using the PyTorch framework on an RTX 3090 GPU.

\renewcommand{\arraystretch}{1.2}

\begin{table}[h]
    \caption{Performance comparison with state-of-the-art HNNs on hypergraph datasets: mean accuracy (\%) ± standard deviation. \textbf{Bold font} indicates the best result, while \underline{best baselines} are underlined.} 
  \label{reordered_hypergraph_experiment}
  \centering
  \resizebox{0.99\linewidth}{!}{ 
  \begin{tabular}{ccccc}
    \toprule
    Methods & Citeseer & Senate & ModelNet40 & NTU2012 \\
    \midrule
    HGNN        & $73.07\% \pm 1.35$  & $59.72 \% \pm 4.61$ & $95.27 \% \pm 0.37$ & $86.74 \% \pm 1.31$ \\
    HyperGCN    & $71.52\% \pm 1.28$  & $52.68 \% \pm 7.72$ & $75.07 \% \pm 3.26$ & $53.76 \% \pm 4.63$ \\
    %LEGCN       & $72.27\% \pm 0.96$  & $\underline{75.04 \% \pm 6.37}$ & $96.42 \% \pm 0.25$ & $87.02 \% \pm 1.68$ \\
    HNHN        & $72.13\% \pm 1.43$  & $53.52 \% \pm 6.04$ & $97.19 \% \pm 0.35$ & $88.65 \% \pm 1.30$ \\
    HCHA        & $72.48\% \pm 1.36$  & $51.63 \% \pm 4.16$ & $94.60 \% \pm 0.29$ & $87.41 \% \pm 1.64$ \\
    UniGNNII    & $73.13\% \pm 1.30$  & $54.23 \% \pm 5.68$ & $97.50 \% \pm 0.31$ & $\underline{88.77 \% \pm 0.89}$ \\
    $\mathrm{HGNN}^{+}$ & $73.02\% \pm 1.43$  & $55.63 \% \pm 6.88$ & $94.35 \% \pm 0.39$ & $86.66 \% \pm 1.33$ \\
    AllSetTransformer & $73.08\% \pm 0.74$  & $53.81 \% \pm 5.30$ & $\underline{98.11 \% \pm 0.26}$ & $88.67 \% \pm 1.29$ \\
    AllDeepSets        & $70.08\% \pm 1.72$  & $56.90 \% \pm 5.22$ & $96.81 \% \pm 0.36$ & $88.39 \% \pm 1.25$  \\
    EDHNN       & $\underline{74.03\% \pm 1.27}$  & $\underline{65.49 \% \pm 2.02}$ & $97.28 \% \pm 0.27$ & $87.95 \% \pm 0.94$ \\

    \midrule
    \textbf{\projectname} & \textbf{75.05\% $\pm$ 1.16}  & \textbf{76.37\% $\pm$ 6.84} & \textbf{98.52\% $\pm$ 0.18} & \textbf{90.58\% $\pm$ 1.48} \\
    \bottomrule
  \end{tabular}
  }
\end{table}

\subsection{Experiment Results}
\textbf{Comparison with SOTA methods.} Table I presents a comparative analysis of our \projectname model against ten hypergraph baselines. The results indicate that, while the baseline models show varying strengths and weaknesses across different datasets, \projectname consistently achieves superior performance, outperforming all state-of-the-art models.
Notably, on the Senate dataset, \projectname surpasses the second-best model by nearly 9\% on average. This substantial gain in accuracy is primarily attributed to two factors. This significant improvement in accuracy is primarily due to two factors. First, our feature extraction and adjustment mechanism effectively balances the feature acquisition capabilities between high-degree and low-degree vertices. 
This approach allows low-degree vertices to capture more meaningful features while preventing high-degree vertices from accumulating redundant information. Second, the Kolmogorov-Arnold Networks (KANs) enhance the model's capacity to capture complex, nonlinear relationships, providing a deeper understanding of high-order associations within hypergraphs. The additional encoding by KAN layers allows our model to better interpret these intricate relationships, leading to higher overall accuracy.

\begin{table}[ht]
\caption{Result of the ablation study.}
  \label{ab_experiment}
	\centering
  \resizebox{0.99\linewidth}{!}{ 
	\begin{tabular}{ccccc}
		\hline
        Dataset& w/o both &  w/o FE\&FA &  w/o KAN & with both \\
        \midrule
		Citeseer& 71.22\% ± 1.16 & 73.50\% ± 1.61    & 73.65\% ± 0.93 & 75.05\% ± 1.16 \\
        % Cora &  80.52± 0.98   & 81.05 ± 1.27 & 81.15 ± 1.30 \\
		NTU2012& 87.80\% ± 2.05 & 88.89\% ± 1.49 & 89.86\% ± 1.91 & 90.58\% $\pm$ 1.48 \\
        % Senate &  75.04± 4.06   & 75.92 ± 4.72 & 77.55 ± 2.65 \\
		\hline
	\end{tabular}
 }
\end{table}

\textbf{Ablation Study.}
We conduct a series of ablation studies to assess the importance of various modules in our \projectname model using two selected datasets. We create two model variants: \projectname (w/o FE\&FA), which excludes the Feature Extraction (FE) and Feature Adjustment (FA) modules, and \projectname (w/o KAN), which omits the KAN layers. Additionally, we consider \projectname (w/o both), which removes all the modules. The results of these ablation studies, presented in Table \ref{ab_experiment}, demonstrate that both \projectname (w/o FE\&FA) and \projectname (w/o KAN) perform worse than the full \projectname model, underscoring the essential role of the FE, FA, and KAN modules. Notably, \projectname (w/o both) shows the poorest performance, highlighting the critical importance of our complete approach in enhancing model discriminability.

\vspace{-1em}
\begin{figure}[ht!]
  \centering
  \includegraphics[width=0.49\textwidth]{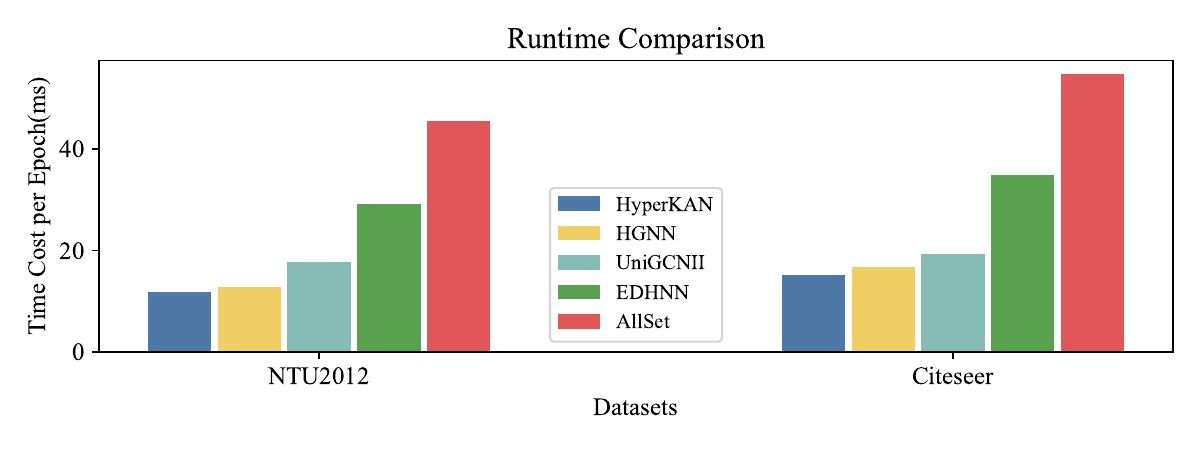}
  \vspace{-2em}
  \caption{Comparison of runtime.}
\label{hypermeter}
\end{figure}
\vspace{-0.5em}
\textbf{Runtime Comparison.} We conducted experiments on the NTU2012 and Citeseer datasets to compare the runtime speed of popular HNN models with \projectname, as shown in Fig.~\ref{hypermeter}.  The results show that \projectname's runtime is comparable to that of HGNN, yet it significantly outperforms HGNN in accuracy. Furthermore, \projectname demonstrates superior performance compared to state-of-the-art methods, including EDHNN, AllSet, and UniGCNII, while also maintaining lower computational costs. This highlights \projectname's efficiency in achieving high accuracy without significantly increasing the total resource consumption.

\section{Conclusion}
In this work, we introduce \projectname, a novel framework for hypergraph representation learning. Our approach begins by extracting essential structural features from the hypergraph, followed by employing Kolmogorov-Arnold Networks (KANs) for deeper encoding and feature extraction. This method enables high-degree vertices to avoid redundant features, while low-degree vertices can effectively acquire sufficient structural information. Extensive experiments on real-world datasets demonstrate that \projectname consistently outperforms state-of-the-art methods, underscoring its effectiveness and efficiency in hypergraph representation learning.

\bibliographystyle{IEEEtran}
\bibliography{main} 

\begin{thebibliography}{10}
\providecommand{\url}[1]{#1}
\csname url@samestyle\endcsname
\providecommand{\newblock}{\relax}
\providecommand{\bibinfo}[2]{#2}
\providecommand{\BIBentrySTDinterwordspacing}{\spaceskip=0pt\relax}
\providecommand{\BIBentryALTinterwordstretchfactor}{4}
\providecommand{\BIBentryALTinterwordspacing}{\spaceskip=\fontdimen2\font plus
\BIBentryALTinterwordstretchfactor\fontdimen3\font minus \fontdimen4\font\relax}
\providecommand{\BIBforeignlanguage}[2]{{%
\expandafter\ifx\csname l@#1\endcsname\relax
\typeout{** WARNING: IEEEtran.bst: No hyphenation pattern has been}%
\typeout{** loaded for the language `#1'. Using the pattern for}%
\typeout{** the default language instead.}%
\else
\language=\csname l@#1\endcsname
\fi
#2}}
\providecommand{\BIBdecl}{\relax}
\BIBdecl

\bibitem{li2013link}
D.~Li, Z.~Xu, S.~Li, and X.~Sun, ``Link prediction in social networks based on hypergraph,'' in \emph{Proceedings of the 22nd international conference on world wide web}, 2013, pp. 41--42.

\bibitem{meng2024link}
C.~Meng and H.~Motevalli, ``Link prediction in social networks using hyper-motif representation on hypergraph,'' \emph{Multimedia Systems}, vol.~30, no.~3, p. 123, 2024.

\bibitem{guan2023sparse}
Y.~Guan, X.~Sun, and Y.~Sun, ``Sparse relation prediction based on hypergraph neural networks in online social networks,'' \emph{World Wide Web}, vol.~26, no.~1, pp. 7--31, 2023.

\bibitem{10887924}
X.~Fang, C.~Huan, B.~Wang, S.~Ma, H.~Zhang, and C.~Zhao, ``Hypersf: A hypergraph representation learning method based on structural fusion,'' in \emph{ICASSP 2025 - 2025 IEEE International Conference on Acoustics, Speech and Signal Processing (ICASSP)}, 2025, pp. 1--5.

\bibitem{feng2021hypergraph}
S.~Feng, E.~Heath, B.~Jefferson, C.~Joslyn, H.~Kvinge, H.~D. Mitchell, B.~Praggastis, A.~J. Eisfeld, A.~C. Sims, L.~B. Thackray \emph{et~al.}, ``Hypergraph models of biological networks to identify genes critical to pathogenic viral response,'' \emph{BMC bioinformatics}, vol.~22, no.~1, p. 287, 2021.

\bibitem{jin2023general}
S.~Jin, Y.~Hong, L.~Zeng, Y.~Jiang, Y.~Lin, L.~Wei, Z.~Yu, X.~Zeng, and X.~Liu, ``A general hypergraph learning algorithm for drug multi-task predictions in micro-to-macro biomedical networks,'' \emph{PLOS Computational Biology}, vol.~19, no.~11, p. e1011597, 2023.

\bibitem{lugo2021classification}
J.~Lugo-Martinez, D.~Zeiberg, T.~Gaudelet, N.~Malod-Dognin, N.~Przulj, and P.~Radivojac, ``Classification in biological networks with hypergraphlet kernels,'' \emph{Bioinformatics}, vol.~37, no.~7, pp. 1000--1007, 2021.

\bibitem{shi2020point}
W.~Shi and R.~Rajkumar, ``Point-gnn: Graph neural network for 3d object detection in a point cloud,'' in \emph{Proceedings of the IEEE/CVF conference on computer vision and pattern recognition}, 2020, pp. 1711--1719.

\bibitem{pradhyumna2021graph}
P.~Pradhyumna, G.~Shreya \emph{et~al.}, ``Graph neural network (gnn) in image and video understanding using deep learning for computer vision applications,'' in \emph{2021 Second International Conference on Electronics and Sustainable Communication Systems (ICESC)}.\hskip 1em plus 0.5em minus 0.4em\relax IEEE, 2021, pp. 1183--1189.

\bibitem{han2022vision}
K.~Han, Y.~Wang, J.~Guo, Y.~Tang, and E.~Wu, ``Vision gnn: An image is worth graph of nodes,'' in \emph{Advances in Neural Information Processing Systems}, 2022.

\bibitem{antelmi2023survey}
A.~Antelmi, G.~Cordasco, M.~Polato, V.~Scarano, C.~Spagnuolo, and D.~Yang, ``A survey on hypergraph representation learning,'' \emph{ACM Computing Surveys}, vol.~56, no.~1, pp. 1--38, 2023.

\bibitem{liu2024hypergraph}
Z.~Liu, B.~Tang, Z.~Ye, X.~Dong, S.~Chen, and Y.~Wang, ``Hypergraph transformer for semi-supervised classification,'' in \emph{ICASSP 2024-2024 IEEE International Conference on Acoustics, Speech and Signal Processing (ICASSP)}.\hskip 1em plus 0.5em minus 0.4em\relax IEEE, 2024, pp. 7515--7519.

\bibitem{kim2024hypeboy}
S.~Kim, S.~Kang, F.~Bu, S.~Y. Lee, J.~Yoo, and K.~Shin, ``Hypeboy: Generative self-supervised representation learning on hypergraphs,'' \emph{arXiv preprint arXiv:2404.00638}, 2024.

\bibitem{huang2021unignn}
J.~Huang and J.~Yang, ``Unignn: a unified framework for graph and hypergraph neural networks,'' in \emph{Proceedings of the Thirtieth International Joint Conference on Artificial Intelligence, IJCAI-21}, 2021.

\bibitem{chien2021you}
E.~Chien, C.~Pan, J.~Peng, and O.~Milenkovic, ``You are allset: A multiset function framework for hypergraph neural networks,'' in \emph{International Conference on Learning Representations}, 2022.

\bibitem{dong2020hnhn}
Y.~Dong, W.~Sawin, and Y.~Bengio, ``Hnhn: Hypergraph networks with hyperedge neurons,'' \emph{arXiv preprint arXiv:2006.12278}, 2020.

\bibitem{yadati2019hypergcn}
N.~Yadati, M.~Nimishakavi, P.~Yadav, V.~Nitin, A.~Louis, and P.~Talukdar, ``Hypergcn: A new method for training graph convolutional networks on hypergraphs,'' \emph{Advances in neural information processing systems}, vol.~32, 2019.

\bibitem{sun2008hypergraph}
L.~Sun, S.~Ji, and J.~Ye, ``Hypergraph spectral learning for multi-label classification,'' in \emph{Proceedings of the 14th ACM SIGKDD international conference on Knowledge discovery and data mining}, 2008, pp. 668--676.

\bibitem{yang2022semi}
C.~Yang, R.~Wang, S.~Yao, and T.~Abdelzaher, ``Semi-supervised hypergraph node classification on hypergraph line expansion,'' in \emph{Proceedings of the 31st ACM International Conference on Information \& Knowledge Management}, 2022, pp. 2352--2361.

\bibitem{yan2024hypergraph}
Y.~Yan, Y.~Chen, S.~Wang, H.~Wu, and R.~Cai, ``Hypergraph joint representation learning for hypervertices and hyperedges via cross expansion,'' in \emph{Proceedings of the AAAI Conference on Artificial Intelligence}, vol.~38, no.~8, 2024, pp. 9232--9240.

\bibitem{feng2019hypergraph}
Y.~Feng, H.~You, Z.~Zhang, R.~Ji, and Y.~Gao, ``Hypergraph neural networks,'' in \emph{Proceedings of the AAAI conference on artificial intelligence}, vol.~33, 2019, pp. 3558--3565.

\bibitem{arya2020hypersage}
D.~Arya, D.~K. Gupta, S.~Rudinac, and M.~Worring, ``Hypersage: Generalizing inductive representation learning on hypergraphs,'' \emph{arXiv preprint arXiv:2010.04558}, 2020.

\bibitem{tudisco2021nonlinear}
F.~Tudisco, K.~Prokopchik, and A.~R. Benson, ``A nonlinear diffusion method for semi-supervised learning on hypergraphs,'' \emph{arXiv preprint arXiv:2103.14867}, 2021.

\bibitem{bai2021hypergraph}
S.~Bai, F.~Zhang, and P.~H. Torr, ``Hypergraph convolution and hypergraph attention,'' \emph{Pattern Recognition}, vol. 110, p. 107637, 2021.

\bibitem{wang2022equivariant}
P.~Wang, S.~Yang, Y.~Liu, Z.~Wang, and P.~Li, ``Equivariant hypergraph diffusion neural operators,'' in \emph{International Conference on Learning Representations (ICLR)}, 2023.

\bibitem{duta2023sheaf}
I.~Duta, G.~Cassarà, F.~Silvestri, and P.~Liò, ``Sheaf hypergraph networks,'' 2023.

\bibitem{liu2024kan}
Z.~Liu, Y.~Wang, S.~Vaidya, F.~Ruehle, J.~Halverson, M.~Solja{\v{c}}i{\'c}, T.~Y. Hou, and M.~Tegmark, ``Kan: Kolmogorov-arnold networks,'' \emph{arXiv preprint arXiv:2404.19756}, 2024.

\bibitem{sa}
Bresson, ``Kagnns: Kolmogorov-arnold networks meet graph learning,'' \emph{arXiv preprint arXiv:2406.18380}, 2024.

\bibitem{de2024kolmogorov}
G.~De~Carlo, A.~Mastropietro, and A.~Anagnostopoulos, ``Kolmogorov-arnold graph neural networks,'' \emph{arXiv preprint arXiv:2406.18354}, 2024.

\bibitem{kiamari2024gkan}
M.~Kiamari, M.~Kiamari, and B.~Krishnamachari, ``Gkan: Graph kolmogorov-arnold networks,'' \emph{arXiv preprint arXiv:2406.06470}, 2024.

\bibitem{fowler2006legislative}
J.~H. Fowler, ``Legislative cosponsorship networks in the us house and senate,'' \emph{Social networks}, vol.~28, no.~4, pp. 454--465, 2006.

\bibitem{chodrow2021hypergraph}
P.~S. Chodrow, N.~Veldt, and A.~R. Benson, ``Hypergraph clustering: from blockmodels to modularity,'' \emph{arXiv e-prints}, pp. arXiv--2101, 2021.

\bibitem{wu20153d}
Z.~Wu, S.~Song, A.~Khosla, F.~Yu, L.~Zhang, X.~Tang, and J.~Xiao, ``3d shapenets: A deep representation for volumetric shapes,'' in \emph{Proceedings of the IEEE conference on computer vision and pattern recognition}, 2015, pp. 1912--1920.

\bibitem{chen2003visual}
D.-Y. Chen, X.-P. Tian, Y.-T. Shen, and M.~Ouhyoung, ``On visual similarity based 3d model retrieval,'' in \emph{Computer graphics forum}, vol.~22, no.~3.\hskip 1em plus 0.5em minus 0.4em\relax Wiley Online Library, 2003, pp. 223--232.

\bibitem{9795251}
Y.~Gao, Y.~Feng, S.~Ji, and R.~Ji, ``Hgnn+: General hypergraph neural networks,'' \emph{IEEE Transactions on Pattern Analysis and Machine Intelligence}, vol.~45, no.~3, pp. 3181--3199, 2023.

\bibitem{kingma2014adam}
D.~P. Kingma and J.~Ba, ``Adam: {A} method for stochastic optimization,'' in \emph{3rd International Conference on Learning Representations, {ICLR} 2015, San Diego, CA, USA, May 7-9, 2015, Conference Track Proceedings}, 2015.

\end{thebibliography}

\end{document}